# Federated Multi-Agent Deep Learning and Neural Networks for Advanced Distributed Sensing in Wireless Networks

Nadine Muller, Stefano DeRosa, Su Zhang, Chun Lee Huan
Faculty of Information Technology and Electrical Engineering, University of Oulu

**Abstract**—Multi-agent deep learning (MADL), including multi-agent deep reinforcement learning (MADRL), distributed/federated training, and graph-structured neural networks, is becoming a unifying framework for decision-making and inference in wireless systems where sensing, communication, and computing are tightly coupled. Recent 5G-Advanced and 6G visions strengthen this coupling through integrated sensing and communication, edge intelligence, open programmable RAN, and non-terrestrial/UAV networking, which create decentralized, partially observed, time-varying, and resource-constrained control problems. This survey synthesizes the state of the art, with emphasis on 2021-2025 research, on MADL for distributed sensing and wireless communications. We present a task-driven taxonomy across (i) learning formulations (Markov games, Dec-POMDPs, CTDE), (ii) neural architectures (GNN-based radio resource management, attention-based policies, hierarchical learning, and over-the-air aggregation), (iii) advanced techniques (federated reinforcement learning, communication-efficient federated deep RL, and serverless edge learning orchestration), and (iv) application domains (MEC offloading with slicing, UAV-enabled heterogeneous networks with power-domain NOMA, intrusion detection in sensor networks, and ISAC-driven perceptive mobile networks). We also provide comparative tables of algorithms, training topologies, and system-level trade-offs in latency, spectral efficiency, energy, privacy, and robustness. Finally, we identify open issues including scalability, non-stationarity, security against poisoning and backdoors, communication overhead, and real-time safety, and outline research directions toward 6G-native sense-communicate-compute-learn systems.

## Introduction

Wireless networks are transitioning from largely communication-centric infrastructures to multi-functional, sensing-capable and compute-rich platforms, motivated by applications such as autonomous mobility, industrial automation, extended reality, and low-altitude/UAV-based services. ISAC (and the broader "perceptive mobile networks" concept) exemplifies this evolution by reusing spectrum, hardware, and signaling to jointly enable communications and environment awareness [1]. These systems are inherently multi-node and multi-perspective: sensing is improved by networked collaboration across many sensing/communication nodes, while communications performance must be preserved under sensing-induced interference and tight latency budgets. [2]

From an optimisation viewpoint, next-generation wireless control involves high-dimensional, stochastic, and coupled decision variables (such as association, power, beamforming, spectrum access, trajectory control, caching, and service placement), often under partial observability and decentralised information. In such settings, multi-agent deep learning provides (i) scalable function approximation for complex policies and value functions, (ii) principled frameworks for decentralised coordination (cooperative or competitive), and (iii) data-driven adaptation when accurate analytical models are unavailable or intractable. A representative tutorial for AI-enabled wireless networks emphasises that multi-agent reinforcement learning is particularly relevant when more than one entity makes decisions, for example, UAVs, access points, slices, or edge servers. [3]

A second, equally important inflection is distributed training and privacy-aware intelligence at the wireless edge. Federated learning (FL) enables collaborative model training without moving raw data, aligning well with privacy constraints and the data locality of IoT/edge sensing. Yet wireless FL must co-design learning and communications (for example, client selection, bandwidth/power allocation, over-the-air aggregation), and it inherits new threats such as model/data poisoning and backdoors, which are particularly acute in open wireless environments. [4]

This survey makes the following contributions:

1) A systems-to-algorithms taxonomy of MADL across distributed sensing and wireless communications, integrating MADRL, FL/FRL, and GNN-based architectures [5].

2) A critical synthesis of advanced techniques, hierarchical/over-the-air FL, communication-efficient federated deep RL, and serverless edge computing as an integrated orchestration substrate [6].

3) Application-focused coverage spanning MEC offloading and slicing, UAV-enabled heterogeneous networks (including power-domain NOMA and federated multi-agent control), intrusion detection in sensing networks, and ISAC-driven perceptive distributed networking [7].

4) An open-issues agenda for 6G-native deployments, emphasising scalability, security, real-time constraints, and reproducibility [8].

# Background and Fundamentals

Multi-agent learning in wireless systems is best understood by aligning wireless primitives (nodes, channels, traffic, mobility, slices, and sensing measurements) with learning primitives (agents, observations, actions, rewards/losses, and coordination mechanisms). A widely used modelling approach is the Markov game (a multi-agent generalization of MDPs) or its partially observed and decentralized variants (such as, Dec-POMDPs), commonly paired with centralized training and decentralized execution (CTDE). This is especially natural for wireless networks: joint training can exploit global simulators or network-level telemetry, while execution must be decentralized to meet signaling and latency constraints. [9]

Distributed sensing technologies and ISAC context. Modern distributed sensing spans conventional wireless sensor networks (WSNs), cooperative localization, multi-static radar-like sensing, and ISAC-enabled perceptive systems. Key surveys emphasize that joint radar/communications (or sensing/communications) integration requires coordinated waveform design, receiver processing, and cross-layer resource management, while ISAC fundamental-limit analyses highlight intrinsic trade-offs between information transfer and sensing performance. [10]

Perceptive mobile networks extend this by treating cellular infrastructure as a networked sensing platform where multiple nodes collaboratively observe targets from different perspectives; this increases sensing robustness but raises challenges such as mutual interference and fast target tracking. [11]

Neural network architectures in wireless learning. While generic multilayer perceptrons can approximate policies, wireless problems exhibit strong structure: interference graphs, neighbour interactions, permutation symmetry across users/links, and locality. Graph neural networks (GNNs) explicitly encode such structure, enabling scalable and generalizable learning-to-optimize for radio resource management (RRM) [12]. In particular, message-passing GNN approaches formalize permutation equivariance for RRM tasks and provide both performance and interpretability advantages, while subsequent work demonstrates broader "from theory to practice" pipelines for wireless communications.

Edge intelligence and 6G framing. Edge AI visions in 6G argue that intelligence must be "pushed down" to edge nodes to meet stringent latency and reliability requirements and to exploit local data. This motivates distributed training (e.g., FL) and distributed inference, and it ties directly to multi-agent control, since learning agents and inference nodes often coincide with network entities that must act in real time. [13]

```
Distributed sensing + wireless comms + MADL (conceptual)
```

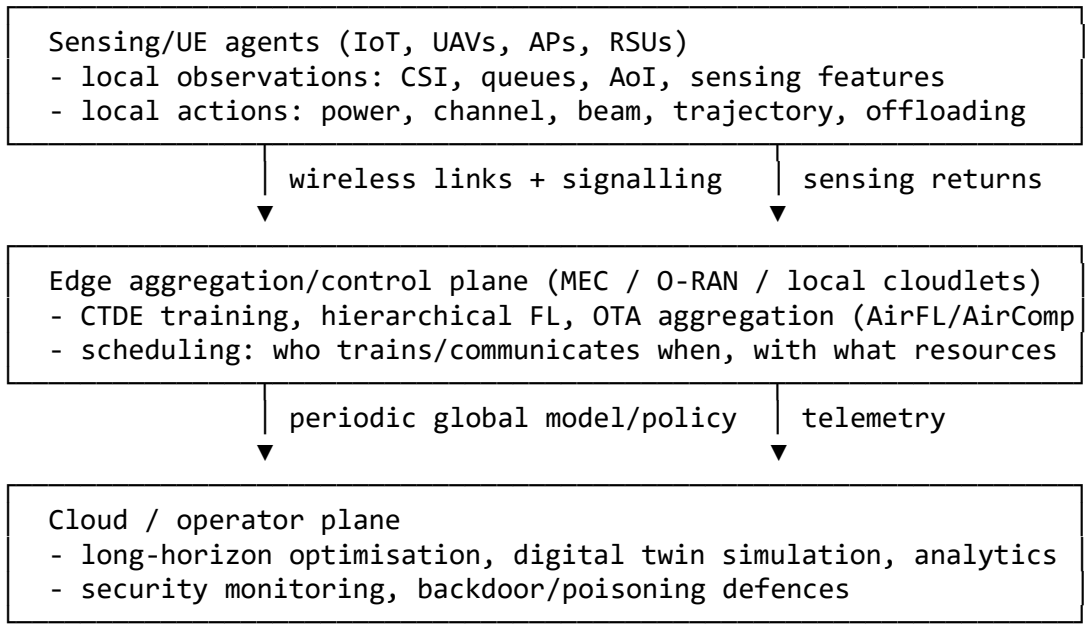

```
   Sensing/UE agents (IoT, UAVs, APs, RSUs)
   - local observations: CSI, queues, AoI, sensing features
   - local actions: power, channel, beam, trajectory, offloading

              │ wireless links + signalling  │ sensing returns
              ▼                              ▼
   Edge aggregation/control plane (MEC / O-RAN / local cloudlets)
   - CTDE training, hierarchical FL, OTA aggregation (AirFL/AirComp
   - scheduling: who trains/communicates when, with what resources

              │ periodic global model/policy │ telemetry
              ▼                              ▼
   Cloud / operator plane
   - long-horizon optimisation, digital twin simulation, analytics
   - security monitoring, backdoor/poisoning defences
```

The remainder of the paper analyses how specific algorithmic families realise this architecture, and where they succeed or fail under wireless-specific constraints. [14]

# Advanced Techniques

This section surveys a set of technique clusters that repeatedly appear in high-impact recent literature: multi-agent DRL, federated (and over-the-air/hierarchical) learning, federated reinforcement learning, and serverless paradigms that decouple control logic from infrastructure.

Multi-agent deep reinforcement learning for wireless control. A consistent pattern in recent TWC/related work is to cast wireless optimisation as multi-agent sequential decision-making: each base station, access point, user, UAV, or slice controller acts as an agent. For example, TWC research demonstrates that multi-agent DRL can achieve strong performance in resource management problems by learning decentralised policies that minimise coordination overhead at run time while still capturing coupling (such as interference) [15].

Recent work expands the algorithmic toolbox beyond basic DQN/DDPG-style methods, incorporating (i) mean-field approximations for scalability in dense networks (useful when the number of interacting neighbours is large), and (ii) constrained or safety-aware formulations when QoS/SLA constraints are non-negotiable. [16]

Graph-based deep learning as scalable inductive bias. GNN-based "learning to optimise" frameworks exploit the graph structure induced by channel/interference relationships and naturally generalise across network sizes and topologies. The JSAC architecture-and-theory paper formalises why permutation equivariance matters for dense networks and connects message passing to distributed optimisation; later TWC work provides practical end-to-end treatment for wireless communication tasks. [12]

Federated learning for wireless networks: from privacy to co-design. Wireless FL literature consistently shows that "training over the air" is not simply classical FL on a different transport. Joint design of learning and communications (for example, user selection, power/bandwidth allocation) is central for convergence and energy. A canonical TWC framework explicitly optimises learning objectives under wireless resource limitations. [17]

Further, hierarchical FL architectures (client–edge–cloud) offer a structured way to reduce communication load and latency; convergence-sensitive quantisation and schedule design become key, as detailed in TWC convergence/system-design analyses [18].

Over-the-air computation (AirComp) and AirFL compress aggregation latency by exploiting waveform superposition, but introduce new error mechanisms (fading/noise, device heterogeneity) and require careful power control and aggregation design; recent surveys and TWC studies formalise these trade-offs [19].

Federated reinforcement learning and federated deep RL. Federated reinforcement learning (FRL) is increasingly used when (i) agents accumulate local trajectories or logs that cannot be centralised, and (ii) policies must adapt to heterogeneous environments. A recent IEEE Open Journal survey consolidates FRL fundamentals, challenges, and future directions for wireless networks [20].

On the systems side, "federated deep RL" has been applied to problems such as cooperative edge caching (multiple fog access points train shared policies with reduced convergence issues) and high-mobility / aerial relay networks (joint trajectory and beamforming variables) [21].

Serverless edge computing as an enabling substrate for distributed learning. Serverless edge computing is increasingly positioned as a way to manage the edge–cloud continuum by abstracting infrastructure and enabling event-driven compute placement. A widely cited IEEE Internet Computing survey evaluates maturity and open challenges for serverless at the edge [22]. From an optimisation perspective, cross-edge orchestration work in IEEE Transactions on Services Computing addresses cold-start/caching dynamics through probabilistic caching and request distribution, which is directly relevant to latency-sensitive learning pipelines deployed as micro-functions [23].

Importantly for this survey's scope, "serverless" can be interpreted not only as compute deployment but also as network intelligence deployment: learning components (policy inference, aggregation, monitoring) can be packaged as ephemeral functions invoked on demand, aligning with variable traffic intensity and heterogeneous device participation—an idea reflected in emerging network and edge-learning discussions [24].

## Comparative table of core learning paradigms and wireless suitability

| Paradigm | Typical training topology | Key strength in wireless | Key limitation in wireless | Representative recent sources |
|---|---|---|---|---|
| CTDE MADRL (Markov games) | Centralised training; decentralised execution | Captures coupling (interference, coordination) while keeping run-time signalling low | Non-stationarity; scaling to many agents; safety/QoS constraints | Feriani & Hossain tutorial; TWC resource mgmt; in-X subnetworks; power-control MARL [25] |
| Mean-field MADRL | CTDE + mean-field approximation | Scales to dense deployments by approximating neighbour influence | Approximation error; may miss rare but critical interactions | Unlicensed spectrum mean-field DRL (TWC) [26] |
| GNN-based learning-to-optimise | Supervised/self-supervised training; distributed inference via message passing | Exploits permutation equivariance; generalises across topologies | Requires graph construction/feature design; robustness to missing CSI | JSAC GNN-RRM; TWC GNN practice; 6G optimisation survey [12] |
| Wireless FL (client–server) | Periodic aggregation; resource-aware scheduling | Privacy by design; leverages local data | Communication bottleneck; system heterogeneity; security threats | Wireless FL frameworks; energy-efficient FL; IRS-assisted FL [27] |
| Hierarchical / OTA FL | Multi-tier aggregation or AirComp | Lower latency aggregation; scalable edge intelligence | Aggregation error; interference/heterogeneity; complex control | Hierarchical FL with quantisation; scalable |

| | | | | hierarchical OTA-FL; OTA aggregation over time-varying channels [28] |
|---|---|---|---|---|
| Federated RL / federated deep RL | Local RL; federated policy/model fusion | Distributed adaptation without sharing trajectories | Instability; partial observability; trust/poisoning risks | FRL survey (IEEE OJVT); federated deep RL in TWC applications [29] |
| Serverless edge orchestration | Event-driven function placement/caching | Elasticity; simplified deployment; fit for intermittent learning tasks | Cold starts; resource fragmentation; orchestration complexity | IEEE Internet Computing survey; IEEE TSC orchestration work [30] |

# Applications

This section organises representative applications by where multi-agent learning enters the stack (edge sensing, RAN control, MEC, aerial networking) and what system objective is optimised (throughput, latency, energy, reliability, fairness, privacy, sensing accuracy).

Mobile edge computing offloading, slicing, and adaptive control. Dynamic computation offloading is a canonical application where local traffic, radio conditions, and edge resource availability evolve quickly. A recent IEEE Communications Letters study proposes dynamic offloading in MEC with traffic-aware network slicing using an adaptive TD3 strategy, explicitly coupling learning with slicing-awareness, illustrating how RL agents must internalise both compute and network virtualisation dynamics [31].

In parallel, the Open RAN (O-RAN) movement expands the feasible action space for learning-based RAN control by exposing programmable interfaces and near-real-time control loops. Constraint-aware multi-agent RL has been used to deliver flexible RAN slicing under varying numbers of slices, addressing over-provisioning/SLA-violation risks in earlier RL slicing methods [32].

Complementarily, IEEE Communications Magazine surveys consolidate DRL methods for slice scaling/placement and RL for RAN slice RRM, providing a bridge between algorithm selection and operational requirements. [33]

UAV-enabled and non-terrestrial networks. UAV-based networking introduces fast, coupled control over mobility and communications. Multi-agent deep RL has been used for UAV trajectory optimisation in differentiated-services settings (TMC), while federated multi-agent DRL in TWC targets fair communications and trajectory control across multiple UAVs [34].

UAV-enabled heterogeneous cellular networks further add multi-objective optimisation and multiple-access design. The user-provided Transactions on Emerging Telecommunications

Technologies article integrates serverless federated learning with power-domain NOMA, highlighting a trend toward combining (i) distributed learning for privacy and scalability, (ii) aerial/heterogeneous infrastructure, and (iii) non-orthogonal access for spectral efficiency [35].

Federated deep RL for high-dimensional aerial/relay optimisation. Recent TWC work on federated DRL-aided space–aerial–terrestrial relay networks jointly optimises UAV trajectory and beamforming variables alongside RIS and RSMA components, using federated learning to keep training local while still coordinating complex control [36].

Distributed sensing and ISAC-driven perceptive networking. ISAC surveys and PMN discussions motivate multi-agent learning in two roles: (i) resource control (e.g., scheduling, waveform/beam allocation) to balance sensing and communications, and (ii) distributed inference where agents fuse information across nodes for detection/tracking. Surveys of ISAC fundamental limits and waveform design, together with JSTSP system overviews, provide the technical foundation for these roles [37].

In particular, collaborative sensing in perceptive mobile networks stresses that network-level sensing benefits require coordination protocols and interference management, which naturally align with cooperative multi-agent formulations [38].

Security and intrusion detection in sensor/wireless networks. Intrusion detection remains a core WSN application; the user-provided TC-IDS work represents an early trust-based IDS design for WSNs. While modern deployments increasingly consider distributed learning (such as, FL) for privacy and cross-domain training, they also face poisoning/backdoor threats, especially relevant when IDS models are collaboratively trained over wireless edge networks. Recent IEEE ComST surveys emphasise these vulnerabilities and defences in wireless FL, providing a concrete research bridge between "distributed sensing security" and "distributed learning security" [39].

Application-to-method mapping (system-level perspective)

| Application domain | Agents (examples) | Typical objective(s) | Common MADL method choices | Representative sources |
|---|---|---|---|---|
| MEC offloading + slicing | UEs, edge servers, slice orchestrator | Latency, energy, SLA satisfaction | Actor–critic DRL; multi-agent RL for coordinated control; constrained RL | TD3-based slicing-aware offloading; O-RAN slicing via constrained multi-agent RL [40] |
| RAN slicing + O-RAN control loops | xApps, RIC controllers, slice agents | Long-term utility, isolation, admission control | Cooperative multi-agent RL; RL slicing surveys | JSAC constrained multi-agent RL; ComMag surveys; NOMS multi-agent slicing/control [41] |
| UAV networking | UAVs, BSs, UEs | Coverage, fairness, throughput, energy | MADRL (CTDE), federated multi-agent DRL | TMC UAV trajectory MADRL; TWC federated multi-UAV control [34] |
| ISAC / PMNs | BSs/RSUs/sensors | Sensing–comms trade-offs, tracking | Cooperative MARL; GNN-based | ISAC and PMN surveys; joint radar– |

| | | accuracy, interference mgmt | inference/resource policies | comms overview [42] |
| --- | --- | --- | --- | --- |
| Cooperative caching / F-RAN | Fog access points | Delay, hit rate, privacy | Communication-efficient federated deep RL | TWC federated deep RL edge caching [43] |
| Wireless FL / AirFL | Edge devices | Convergence time, energy, spectral efficiency, privacy | Joint learning–comms optimisation; hierarchical/OTA aggregation | TWC joint learning–comms FL; hierarchical FL with quantisation; scalable hierarchical OTA-FL [44] |
| WSN intrusion detection | Sensor nodes, cluster heads | Detection accuracy, false alarm control, energy | Local DL + FL; robust aggregation | TC-IDS; wireless FL security/backdoor survey [45] |

# Challenges and Open Issues

Despite rapid progress, deploying multi-agent deep learning in distributed sensing and wireless communications remains constrained by several systemic and theoretical challenges.

Scalability and non-stationarity. In many wireless settings, the number of controllable entities can be large (dense small cells, massive IoT, UAV swarms). Multi-agent RL faces non-stationarity because each agent's policy update changes the environment seen by others. Mean-field approximations can mitigate scaling issues but introduce modelling error and may underrepresent rare but important interactions (for example, edge-case interference) [46].

Communication overhead and learning–control coupling. Distributed learning is fundamentally limited by the wireless medium it relies on. FL and FRL must budget spectrum and power for model updates while still serving user traffic; over-the-air computation reduces latency but introduces aggregation error and interference sensitivity. Surveys and TWC studies show that convergence and performance are inseparable from communication design (power control, scheduling, quantisation, hierarchical aggregation intervals) [47].

Real-time constraints and safety. Many target use cases, URLLC control, UAV mobility, and ISAC tracking, impose tight deadlines and safety constraints. Constraint-aware RL for RAN slicing illustrates progress toward SLA-aware policies, yet formal safety guarantees and verification remain limited in most deep RL pipelines [48].

Security and privacy under adversarial conditions. Wireless FL is exposed to both statistical heterogeneity and adversarial manipulation. Backdoor and poisoning attacks can be mounted through corrupted clients or manipulated updates; a recent IEEE ComST comprehensive survey Organises attack surfaces and defence mechanisms for wireless FL [49].

Additionally, distributed RL and multi-agent systems can be attacked through observation spoofing, reward manipulation, and byzantine behaviours. While FRL surveys identify challenges and future directions, robust-by-design multi-agent learning for wireless remains an open research frontier [50].

System heterogeneity and reproducibility. Wireless learning experiments often rely on custom simulators, proprietary traces, or narrow parameter regimes. Although high-level surveys and tutorials offer unifying perspectives, reproducible benchmarking across (i) radio/traffic models, (ii) sensing modalities, and (iii) compute/energy constraints is still limited, particularly for joint ISAC + learning studies [51].

Operational deployment complexity (O-RAN and serverless). Programmable O-RAN control loops enable learning-based control but also create operational risk: signalling storms, control-loop instability, and misconfiguration can cascade across network slices and agents. Meanwhile, serverless edge computing can simplify deployment but introduces cold-start and orchestration challenges that may be incompatible with hard real-time objectives unless carefully engineered [52].

# Future Directions

The next phase of research should be driven by 6G-native integration, not only integrating communications and sensing (ISAC), but also tightly integrating communication, sensing, computation, and learning into a single co-designed stack.

Sense–communicate–compute co-design with learning in the loop. ISAC surveys indicate that future systems will require joint optimisation over waveforms, beams, and sensing/communication metrics; multi-agent learning can serve as an adaptive control layer when modelling complexity grows. A key future direction is to build hybrid model-based + learning-based controllers that retain interpretability and constraint handling while gaining adaptation speed [53].

Constraint-aware and verifiable multi-agent policies. Constraint-aware multi-agent RL for RAN slicing is an early example of aligning learning objectives with SLA constraints. Extending this to safety-critical UAV/ISAC and industrial settings likely requires: (i) constrained RL with explicit feasibility layers, (ii) robust training under distribution shift, and (iii) verification/calibration techniques appropriate for neural policies [54].

Graph-native and permutation-equivariant multi-agent intelligence. GNN-based architectures have shown strong promise for scalable RRM due to structural inductive bias and generalisation. A forward-looking theme is graph-native multi-agent learning, where communication among agents and policy representations are unified by graph message passing, potentially enabling consistent scaling across dense terrestrial networks and dynamic aerial meshes [12].

Federated reinforcement learning and communication-efficient federated deep RL at scale. FRL surveys and TWC applications suggest that federated policy learning will become more central as privacy regulations tighten and as data becomes more distributed (for example, across UAV swarms, vehicles, industrial sensors). Research priorities include robust aggregation under byzantine clients, exploration under heterogeneity, and joint optimisation of policy performance vs. communication cost [55].

Serverless learning control planes for elastic edge intelligence. Serverless edge computing surveys argue that abstraction and elasticity are key for managing the edge–cloud continuum. For wireless learning systems, the compelling direction is to treat learning as a set of event-driven, composable services (policy inference, aggregation, monitoring, anomaly detection), deployed near where data is generated and adapted to fluctuating workloads. However, research must address cold starts, orchestration under wireless bandwidth constraints, and integration with O-RAN control loops [56].

Toward multi-agent, privacy-aware, real-time ISAC intelligence. Combining the ISAC stack with FL/FRL opens a pathway to privacy-aware distributed sensing analytics, e.g., collaborative target recognition without centralising raw sensing streams. Achieving this requires new cross-layer protocols that jointly schedule sensing, communications, and learning updates under strict latency constraints and adversarial risk [57].